\documentclass{svproc}
\pdfoutput=1
\usepackage{comment}
\usepackage[utf8]{inputenc} 
\usepackage[T1]{fontenc}    
\usepackage{hyperref}       
\usepackage{url}            
\usepackage{booktabs}       
\usepackage{amsfonts}       
\usepackage{nicefrac}       
\usepackage{microtype}      
\usepackage{lipsum}
\usepackage{footmisc}
\usepackage{multicol}
\usepackage{slashbox}
\usepackage[libertine]{newtxmath}
\usepackage{color}
\usepackage{setspace}
\usepackage{xcolor}
\usepackage{pdflscape}
\usepackage{graphicx} 
\usepackage{float}
\usepackage{algorithm}
\usepackage{algorithmicx}
\usepackage{algpseudocode}	
\usepackage{indentfirst}
\usepackage{xltabular}
\usepackage{array, makecell} %
\usepackage{xcolor}
\usepackage{tikz}
\usepackage{subcaption}

\definecolor{bblue}{HTML}{4F81BD}
\definecolor{rred}{HTML}{C0504D}
\definecolor{ggreen}{HTML}{9BBB59}
\definecolor{ppurple}{HTML}{9F4C7C}
\definecolor{yyellow}{HTML}{FFD700}
\definecolor{ppink}{HTML}{FE6F5E}
\definecolor{purpule}{HTML}{BF94E4}

\title{ Search Algorithms for Automated Hyper-Parameter Tuning} 
\titlerunning{Search Algorithms for Automated Hyper-Parameter Tuning}

{\small \author{
 Leila Zahedi \inst{1,2}, Farid Ghareh Mohammadi\inst{3},  \\Shabnam Rezapour\inst{4}, Matthew W. Ohland\inst{5}, M. Hadi Amini \inst{1,2}  }
\authorrunning{Leila Zahedi et al.}
    \institute{ 
  Knight Foundation School of Computing and Information Sciences  \\ Florida International University (FIU), Miami, FL 33199\\
  \and Sustainability, Optimization, and Learning for InterDependent networks laboratory (solid lab), FIU, Miami, FL 33199
  \and Department of Computer Science, University of Georgia, Athens, Georgia, 30602,\\
  \and Enterprise and Logistics Engineering, FIU, FL 33174\\
  \and School of Engineering Education, Purdue University, West Lafayette, IN, 47907
   \email{lzahe001@cs.fiu.edu, farid.ghm@uga.edu,\\ srezapou@fiu.edu, ohland@purdue.edu, amini@cs.fiu.edu}
  }}

\begin{document}
\maketitle
\begin{abstract}
 Machine learning is a powerful method for modeling in different fields such as education. Its capability to accurately predict students' success makes it an ideal tool for decision-making tasks related to higher education. The accuracy of machine learning models depends on selecting the proper hyper-parameters. However, it is not an easy task because it requires time and expertise to tune the hyper-parameters to fit the machine learning model. In this paper, we examine the effectiveness of automated hyper-parameter tuning techniques to the realm of students' success. Therefore, we develop two automated Hyper-Parameter Optimization methods, namely grid search and random search, to assess and improve a previous study's performance. 
The experiment results show that applying random search and grid search on machine learning algorithms improves accuracy. We empirically show automated methods' superiority on a real-world educational data (MIDFIELD) for tuning HPs of conventional machine learning classifiers. This work emphasizes the effectiveness of automated hyper-parameter optimization while applying machine learning in the education field to aid faculties, directors', or non-expert users' decisions to improve students' success.
\end{abstract}

\section{Introduction}
\parindent0pt$\diamond$\textbf{ Motivation}
The usage of information technologies in various fields has led to increasingly large-scale data derived from multiple settings. One of these settings is education, concerning better understanding students' pathways and the environments they learn in. Educational Data Mining (EDM) is an emerging field focusing on mining datasets to answer educational research questions \cite{fernandes2019educational}\cite{pena2014educational}. One of the crucial EDM applications is predicting students' success \cite{pena2014educational}. There are a variety of approaches to measure students' success, such as graduation rates, on-time completion, or GPA \cite{dominguez2013gamifying}\cite{boumi2019application}\cite{zahedi2020leveraging}. However, these metrics may overestimate or underestimate a sub-population persistence, such as non-traditional students, part-time students, or transferred students. Therefore, these populations are usually neglected in many studies, which leads to a lack of understanding of the educational pathways. This matter behooves the researchers to look for fair metrics to assess students' success. Stickiness is one of these metrics \cite{ohland2012introducing}.\\
Tuning HPs is an essential task to make a predictive model that performs at its best \cite{mohammadi2019parameter}\cite{anelli2019discriminative}. Building such models is the primary goal of ML models \cite{elshawi2019automated}. However, despite HPs' critical role in the resulting predictive models' quality, they have no clear agreeable defaults in different applications. Manually tuning the HPs not only needs a deep understanding of the models but is also impractical, time and cost-inefficient. Hence, it has become vital to automate the process of optimizing the HPs. In Hyper-Parameter Optimization (HPO), we usually aim to use the value of parameters that significantly contribute to improving a model's accuracy. Therefore the search algorithm looks through different combinations of HP configurations (search space), enabling the model to generate the best model among the candidates through an iterative process \cite{yang2020hyperparameter,diaz2017effective}. 
ML models' HPs need to be selected automatically and carefully to be effective in the application. For a new dataset, the optimal parameter values depend on the application and, more specifically, the dataset itself. This work helps educational researchers and institutions better understand and develop ML models by identifying the appropriate set of HPs in an effective way.

\vspace{7pt}$\noindent\diamond$\textbf{ Contribution}
This study applies two automated HPO methods to predict students' graduation accurately to address the issues mentioned above regarding manual tuning. Grid Search (GS) and Random Search (RS) are among automated parameter optimization methods. The advantage of using GS and RS is higher learning accuracy and its capability for parallelization; that is not an option for all the HPO methods. We apply GS and RS to find the most appropriate HPs for different conventional ML algorithms. The reason for selecting various ML algorithms is that the performance of a given ML model not only depends on the fundamental quality of the algorithm but also on the details of its tuning. Therefore, it is hard to decide if a given ML model is genuinely better or better tuned. In this study, leveraging different numbers of ML models provides us with the option to choose the model with the best performance among other ML models rather than exploring a single model.\\
The main contributions are expanding upon existing research by incorporating automated HPO techniques leveraging conventional ML models into a single pipeline to improve prediction performance and enhance educational decision-making. Also, this is the first time HPO is being applied on this dataset (MIDFIELD\cite{ohland2004creation}). Unlike other studies that study short time-spans and single institution datasets, this study dataset considers a 30-year longitudinal dataset for undergraduate students from 16 universities. In this work, we use a modified definition of graduation, stickiness (the fraction that "stick" to the program or persist), for students who came into contact with their programs \cite{asee_peer_36110}, to include the populations that are overestimated or underestimated in previous studies.
This study aims to improve ML models' classification for the MIDFIELD dataset using GS and RS then compare it to the previous work.

In the following sections, we first develop a clear and formal definition of HPO, and we provide a basic understanding of the concepts and methodologies applied. From there, we discuss the implementation and evaluation of these approaches. Next, we cover experimental results, conclusions, and future work ideas.  
\section{Related Work}
$\noindent \diamond$\textbf{ Hyper-parameter Optimization}
Every ML model has some HPs, and tuning them is essential for making a model work at its best. The notion of HPs is different from parameters. HPs are the parameters that need initialization before training the model since they represent the model architecture. In contrast, parameters can get initialized and updated during the model training \cite{kuhn2013applied}.
Automatically tuning the HPs is one of the main tasks in automated ML (AutoML) to release the burden of manual tasks and improve the performance, reproducibility, and fairness of various studies \cite{feurer2019hyperparameter}. There are several optimization techniques for HPO problems. The most common and conventional HPO methods are manual search or grad student descent (GSD), GS, and RS. Each of them is defined as below:\\
\textbf{Manual Search:}
GSD is the most basic HP tuning method and a typical approach among researchers and students\cite{abreu2019automated}. In this method, users try different HP values based on guessing or domain knowledge and repeat this process until they obtain a satisfactory improved result or when they are out of time. GSD needs a deep understanding of the ML models or sufficient time to get good results. However, the complex nature of ML models and large search spaces make it an impractical approach \cite{olof2018comparative}.
Mohammadi \emph{et al.} explored the effect of manual parameter tuning on performance by combining different embedded parameters and improved the accuracy of semantic auto-encoder in an image classification problem \cite{mohammadi2019parameter}.\\
\textbf{Grid Search:}
GS is the brute-force way of searching HPs \cite{claesen2014easy}, with defined lower and higher bound along with specific steps \cite{syarif2016svm}. GS work based on the Cartesian product of the different set of values, evaluate every configuration and return the combination with the best performance \cite{hutter2014automl}. GS has a simple implementation; however, it can be very inefficient for large search spaces due to its exhaustive nature. This problem exacerbates as data dimensionality increases.\\ 
\textbf{Random Search:}
RS is another common and standard method of searching HPs \cite{bergstra2012random}. RS chooses the HPs configurations on a random basis (instead of evaluating every configuration) and repeats this process until the defined resources are over. RS is much faster than GS but does not follow a path to find the optimal configuration. 

\section{Experimental Setup}
\begin{figure}
	\centering
		\includegraphics[width =1 \textwidth]{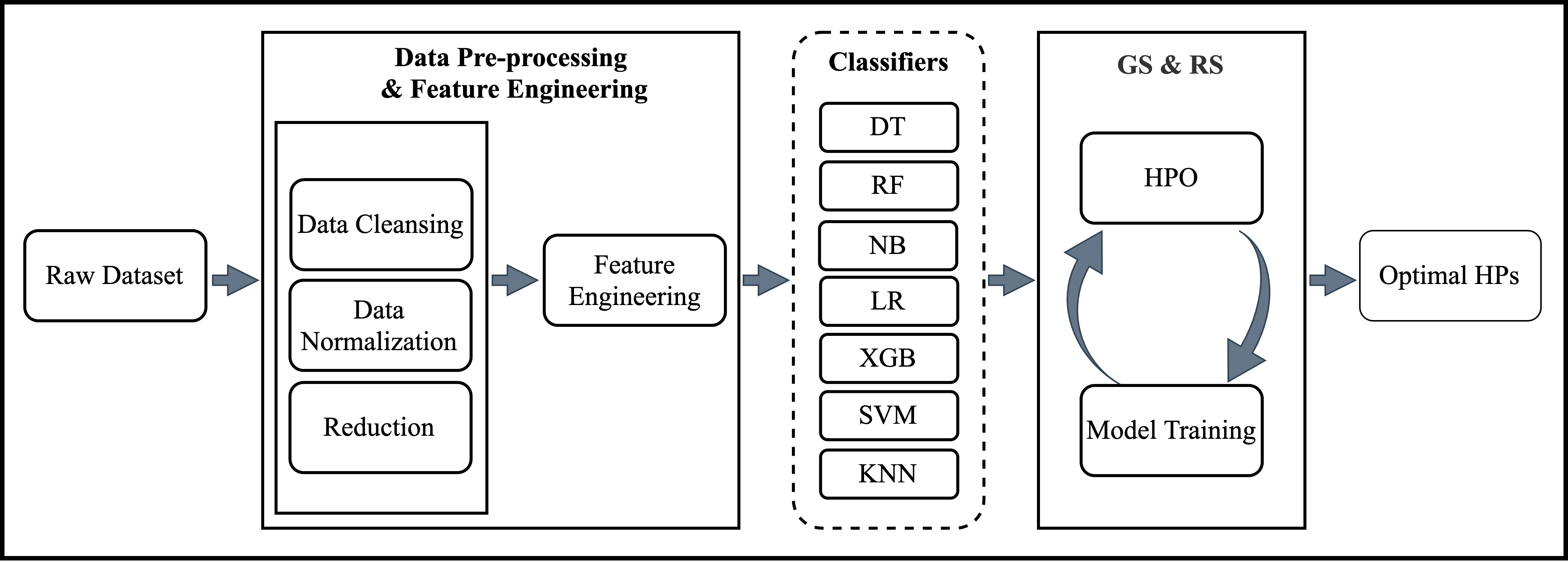}
	\centering
	\caption{Flowchart of the research methodology}
	\label{fig:Method}
\end{figure}

The baseline ML algorithms (model with default parameters), GS, and RS are implemented in this work. 
Figure \ref{fig:Method} shows the research methodology's flowchart. The first phase to conduct is collecting and preparing the data for ML algorithms. Data preparation includes separate tasks such as data reduction, data cleansing, normalization, and feature engineering. 
Data reduction, in this study, refers to using a subset of the dataset rather than the entire dataset. Filtering and sampling are among the most common ways of data reduction. Since we are only interested in students of computing programs, we filter the data and only include the students who enrolled at some point in one of the following majors: computer science, computer engineering, software engineering, computer programming, information technology, and computing and information sciences (CIP=11). In the data cleaning step, which is a step for removing the corrupt and not applicable data, we remove the features with more than 60\% missing values. We also normalized or standardized the range of features of data. After the preprocessing step is the feature engineering step in which we create some new features from raw features that we think might be useful for the power of prediction. Additionally, in this step, we apply one-hot encoding (setting up dummy variables for encoding categorical data) so that we can feed them into ML algorithms. Next, we feed the data to different ML models, including, Decision Tree (DT), Random Forest (RF), Naive Bayes (NB), Logistic Regression (LR), XGBoost (XGB), Support Vector Machine (SVM), and K-Nearest Neighbor(KNN). These models operate as a black box, and therefore, no additional information about building them is required.\\
The next step is the automated HPO method which is an iterative process of choosing HPs. The most commonly used performance metric, accuracy,  is used in our experiments as the classifier performance. Accuracy is the proportion of correctly classified data. We also recorded the tuning time for further considerations. We leveraged GS and RS for finding the optimal set of HPs. For each experiment on the dataset, shuffle 3-fold cross-validation is applied in the training process (to prevent overfitting). Then, we test the model using testing data. In each step, the accuracy of the model is calculated.
Algorithm \ref{alg:GRS} explains GRS pseudo-code starting with initializing HPs and calling RS and GS applied on models and ends with returning the final model, optimal HPs, and accuracy. In this study, the steps defined for GS were defined as 0.5 for continuous values and 1 for all discrete values except \textit {n\_estimator} hyper-parameter in RF and XGB with steps of 5.

\begin{algorithm}
	\caption{Implementation of GRS-AutoHP}
	\begin{algorithmic}[1]
		\Statex \textbf{Input:} \(List_{ML}\) of models (DT, RF, NB, LR, XGB, SVM, KNN) \& raw dataset
        \Statex \textbf{Output:} Best model along performance and optimal architecture 
        
        \State Call PreProcessing
        \For {every model in the \(List_{ML}\) }
		\State Call GS  and RS \Comment{Calculate Accuracy(Acc), HPs}
	
		\If {GS.Acc  larger than RS.Acc}
		\State  Replace model, GS BestHP  
	\EndIf
    \If{RS$.$Acc  larger than RS.Acc}
		\State Replace model, RS BestHP 
	\EndIf
		\EndFor
		\State {Return FinalModel, BestHP, Final.Acc} 
	\end{algorithmic} 
	\label{alg:GRS}
\end{algorithm}

\vspace{5pt}$\noindent\diamond$\textbf{ Dataset} All of our experiments are conducted on MIDFIELD (Multiple-Institution Database for Investigating Engineering Longitudinal Development) \cite{ohland2004creation} to provide practical examples. MIDFIELD is a longitudinal student unit-record dataset from 1988-2018 for all undergraduate, degree-seeking students at partner institutions. MIDFIELD includes everything that appears in students' academic records, such as demographic data (ex. sex, age, and race/ethnicity) and information about major concentration, enrollment, graduation, school and pre-school students' performance. As previous research shows, computing majors have different patterns from other STEM majors\cite{asee_peer_36110}, the data examined in this paper is exclusive to computing fields, with about 45k observations. MIDFIELD is used as a binary-classification problem to predict computing students' success, more specifically graduation. For our classification models, accuracy is used as the classifier performance metric. After completing each experiment, the model with the optimal ML architecture will be returned.\\
In this study, we compare the HPO methods (GS and RS) with the baseline is the ML model with default HPs. Since for each ML model, only a few HPs have significant impacts on the model's performance \cite{yang2020hyperparameter}, in this study, we consider the main HPs of the ML classifiers for automated tuning.
\section{Experimental Results}
$\diamond$\textbf{ Classification Results}
This section discusses the results of our experiments. As mentioned earlier, our first scenario is applying different ML algorithms on the data with their default HP values as our baseline method. The first column in Table \ref{tab:long} shows the accuracy for the baseline. As can be seen in the highlighted areas of the table, results indicate that RF performs better than other ML classifiers and, as a result, the final model in the baseline model selection step. Next, we implement the GS and RS on the same dataset. Columns two and three in Table \ref{tab:long} are the results for these two HPO methods, respectively.
\begin{xltabular}{\textwidth}{|c|c|c|c|c|c|}
\caption{Comparison of baseline, previous work, extended work, GS and RS (\%)}
\label{tab:long}\\
\hline 
\multicolumn{1}{|c|}{\textbf{Classifier}} &
\multicolumn{1}{|c|}{\textbf{Baseline}} &
\multicolumn{1}{c|}{\textbf{Work[5]}} &
\multicolumn{1}{c|}{\textbf{Work[5] Extended}} &
\multicolumn{1}{c|}{\textbf{GS}} &
\multicolumn{1}{c|}{\textbf{RS}} \\ \hline 
\endfirsthead

\multicolumn{3}{c}%
{\tablename\ \thetable{} -- continued from previous page} \\
\hline 
\multicolumn{1}{|c|}{\textbf{Classifier}} &
\multicolumn{1}{|c|}{\textbf{Baseline}} &
\multicolumn{1}{c|}{\textbf{Work[5]}} &
\multicolumn{1}{c|}{\textbf{Work[5] Extended}} &
\multicolumn{1}{c|}{\textbf{GS}} &
\multicolumn{1}{c|}{\textbf{RS}} \\ \hline 
\endhead
\hline \multicolumn{4}{|r|}{{Continued on next page}} \\ \hline
\endfoot
\hline
\endlastfoot
\makecell{NB} & \makecell{69.09} & \makecell{82.25} & \makecell{69.09}
& \makecell{70.49} & \makecell{70.49}\\
\makecell{LR} & \makecell{82.92} & \makecell{83.18} & \makecell{82.92}
& \makecell{83.89} & \makecell{83.86}\\
\makecell{KNN} & \makecell{79.66} & \makecell{75.38} & \makecell{81.43}
& \makecell{84.89} & \makecell{84.89}\\
\makecell{SVM} & \makecell{84.45} & \makecell{85.27} & \makecell{85.06}
& \makecell{87.99} & \makecell{87.43} \\
\makecell{DT} & \makecell{80.59}  & \makecell{86.78} & \makecell{82.08}
& \makecell{87.72} & \makecell{87.45}\\
\makecell{RF} & \makecell{\colorbox{gray!30}{\textbf{85.24}}} &
\makecell{\colorbox{gray!30}{\textbf{88.27}}} & \makecell{\colorbox{gray!30}{\textbf{85.30}}} & \makecell{\colorbox{gray!30}{\textbf{88.34}}} & \makecell{88.37}  \\
\makecell{XGB} & \makecell{85.16} &
\makecell{74.58} & \makecell{85.16} & \makecell{\colorbox{gray!30}{\textbf{88.33}}} & \makecell{\colorbox{gray!30}{\textbf{88.80}}}\\
\end{xltabular}

\begin{figure}%
    \centering
    \subfloat[\centering Comparison with baseline approach]{{\includegraphics[width=5.6cm]{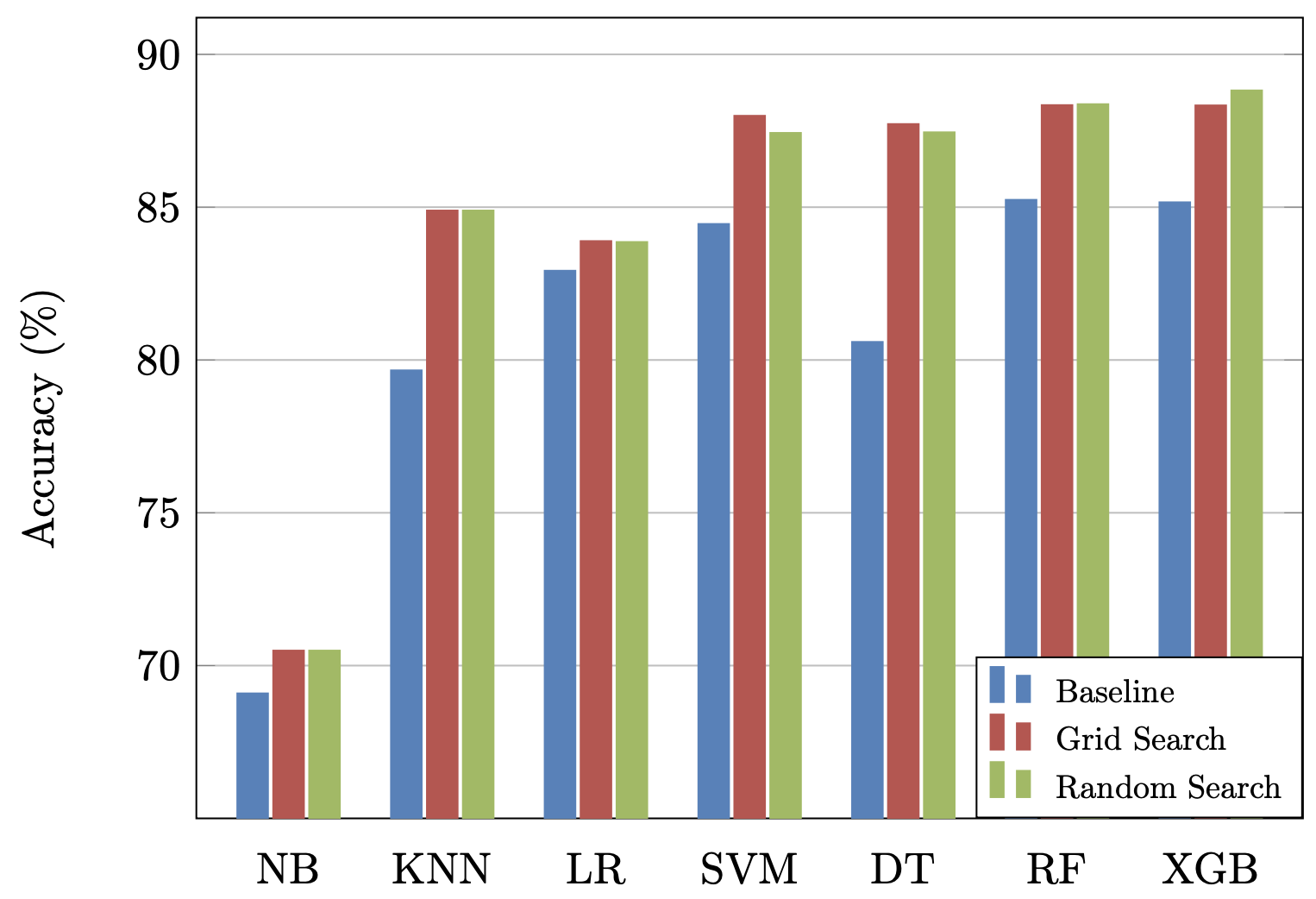} }}%
    \qquad
    \subfloat[\centering Comparison with manual tuning]{{\includegraphics[width=5.6cm]{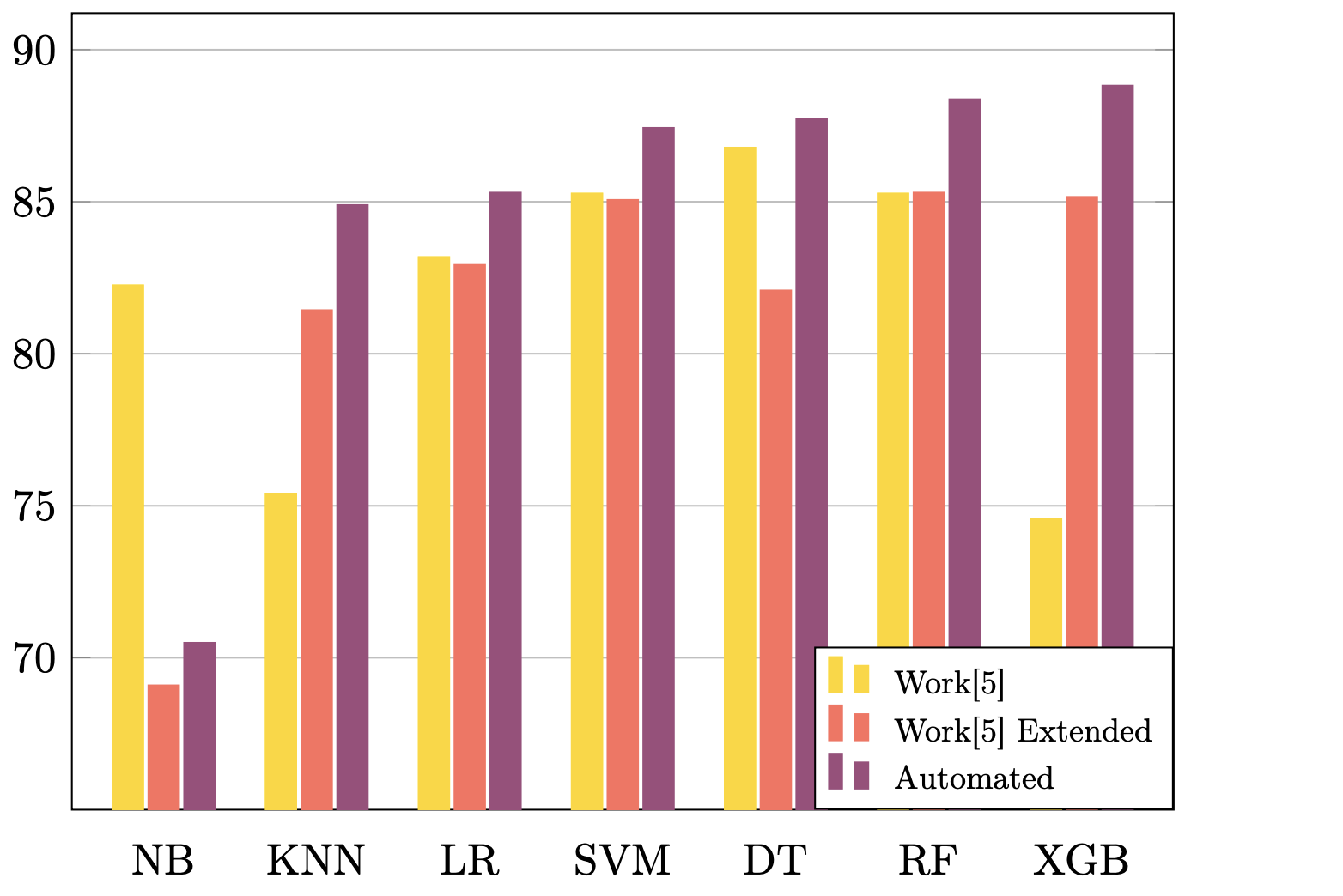} }}%
    \caption{Comparison of automated HPO with manual and baseline approaches}%
    \label{fig:Comparison}%
\end{figure}

The experiment results show that, regardless of the ML model, both automated HPO methods have increased the model performance. However, some models perform better than the rest. For example, NB is not a high-performing model compared to the other models. The reason is that in NB, the probability of each class given different input values requires to be calculated, and no coefficients need to be fitted by the optimization process. However, this characteristic makes the NB faster than other models. Therefore, in cases users want to use NB for their specific purpose, it can be a fast model for HP tuning (NB also has a lower number of HPs).
From Table \ref{tab:long} we can see that after applying GS, XGB is also one of the well-performed candidates among the classifiers, while this is not the case for the baseline approach. As for the RS method, XGB is a well-performed classifier and, as a result, is the final model. This indicates that relying on the default parameters to find the final model is not always the best decision.\vspace{5pt}\\
$\diamond$\textbf{ Comparison with Previous Work }
In this section, we compare the classification results from our experiment with previous work. In the work \cite{zahedi2020leveraging}, an experiment is conducted to predict students' graduation with manual tuning using MIDFIELD (N=39k). In this study, we extended the work\cite{zahedi2020leveraging} and built models using the same configurations (N=45k).\\
As can be seen from the results, automated HPO beats the manual tuning even when the domain knowledge is used to tune the HPs. Automated HPO might take longer than manual tuning, but the results are promising and can guarantee performances close to the global maximum performance.\\ A summary of results for the experiments in this study is shown in Figure \ref{fig:Comparison}. 

\section{Conclusions}
GS and RS methods are applied on seven conventional ML models (DR, NB, SVM, XGB, KNN, LR, and RF) on the MIDFIELD. We use a subset of computing major students from 16 institutions across the U.S. Results show that regardless of the model leveraged, GS and RS improve the classification accuracy. Also, we see that such methods beat the manual tuning methods even when domain knowledge is used. 
Additionally, we observe that using HPO methods, the final selected model is XGB. However, both previous work \cite{zahedi2020leveraging} and our extended work show that RF is the well-performed method. This indicates the importance of automated HPO. We conclude that applying HPO improves the performance and helps to find the proper final model and HPs in the education field.

\section{Discussion and Future Work}
We demonstrated that the GS and RS methods exhibit improved prediction performance. However, these techniques spend a lot of time searching non-promising areas, leading to high tuning time. This is exacerbated when the scale of data or search space increases. Hence, such problems must be solved in a computationally efficient way to have real-time and intelligent decision-making. In our case, using large search spaces and one-hot encoding the features resulted in relatively high tuning time (especially in tree-based models and SVM). High time complexity is an issue that needs future research. Data reduction techniques using nature-inspired algorithms are being used in different applications to achieve reasonable time complexities \cite{mohammadi2020evolutionary}. To achieve this goal, evolutionary algorithms (EAs) have been used widely. One of the EA applications is feature selection to reduce datasets' dimensions without decreasing the performance. Mohammadi \emph{et al.} noted that such algorithms are robust enough to be used in different applications \cite{mohammadi2020applications}.
Going forward, we plan to examine further the impacts of using EA algorithms in educational fields to optimize the ML models' HPs.
\section{Acknowledgment}
All the authors would like to thank Dr. Monique Ross, Florida International University, for her inputs on deploying machine learning for \textit{education research}.

\bibliographystyle{unsrt}  
\bibliography{references}
\end{document}